\pdfoutput=1

\documentclass[11pt]{article}

\usepackage[preprint]{acl}

\usepackage{times}
\usepackage{latexsym}

\usepackage[T1]{fontenc}

\usepackage[utf8]{inputenc}

\usepackage{microtype}

\usepackage{inconsolata}

\usepackage{graphicx}

\usepackage{hyperref}       
\usepackage{url}            
\usepackage{booktabs}       
\usepackage{amsfonts}       
\usepackage{nicefrac}       
\usepackage{xcolor}         

\usepackage{amsmath}
\usepackage{amssymb}
\usepackage{mathtools}
\usepackage{amsthm}

\usepackage{helvet}
\usepackage{courier}
\usepackage{caption}
\usepackage{color}
\usepackage{epsfig}
\usepackage{bm}
\usepackage{makecell}
\usepackage{multirow}
\usepackage{algorithm}
\usepackage{algorithmic}
\usepackage{adjustbox}
\usepackage{bbm}
\usepackage{epstopdf}
\usepackage{threeparttable}
\usepackage{tablefootnote}
\usepackage[normalem]{ulem}
\usepackage{soul}

\usepackage{comment}
\usepackage{subcaption}

\DeclareMathAlphabet{\mathcal}{OMS}{cmsy}{m}{n}

\title{MOOSComp: Improving Lightweight Long-Context Compressor via Mitigating Over-Smoothing and Incorporating Outlier Scores}

\author{
  Fengwei Zhou\thanks{Equal Contribution},
  Jiafei Song\footnotemark[1],
  Wenjin Jason Li\footnotemark[1],
  Gengjian Xue,
  Zhikang Zhao, \\
  {\bf Yichao Lu,
  Bailin Na\thanks{Corresponding Author}} \\
  \small{OPPO CTG} \\
  \small{\{zhoufengwei, songjiafei, liwenjin1, xuegengjian, zhaozhikang, yichao.lu, nabailin\}@oppo.com}
}

\begin{document}
\maketitle
\begin{abstract}
Recent advances in large language models have significantly improved their ability to process long-context input, but practical applications are challenged by increased inference time and resource consumption, particularly in resource-constrained environments. To address these challenges, we propose MOOSComp, a token-classification-based long-context compression method that enhances the performance of a BERT-based compressor by mitigating the over-smoothing problem and incorporating outlier scores. In the training phase, we add an inter-class cosine similarity loss term to penalize excessively similar token representations, thereby improving the token classification accuracy. During the compression phase, we introduce outlier scores to preserve rare but critical tokens that are prone to be discarded in task-agnostic compression. These scores are integrated with the classifier's output, making the compressor more generalizable to various tasks. Superior performance is achieved at various compression ratios on long-context understanding and reasoning benchmarks. Moreover, our method obtains a speedup of 3.3x at a 4x compression ratio on a resource-constrained mobile device.
\end{abstract}

\section{Introduction}

Recent advances in Large Language Models (LLMs) have greatly enhanced their capability to process long-context inputs~\citep{chen2023extending,fu2024data,zhu2024pose,peng2024yarn,han2024lm}, thereby achieving notable progress in natural language understanding and generation~\citep{achiam2023gpt,team2024gemini,dubey2024llama,yang2024qwen2,liu2024deepseek}.
Nonetheless, with ever-growing context lengths, critical challenges must be addressed for LLMs in practice, especially in resource-constrained environments. One primary concern is the prolonged inference time~\citep{zhou2024survey,qin2024mooncake}, particularly pronounced on edge devices~\citep{yao2024minicpm} that permit only limited computational resources. 
Additionally, the resource consumption (for example, memory and processing power) of LLMs increases sharply with longer contexts, resulting in high costs and reduced efficiency in scenarios necessitating real-time deployment. Consequently, techniques for compressing long-context representations are becoming essential to enable more scalable applications.

\begin{figure*}[t]
  \centering
  \includegraphics[width=0.95\textwidth]{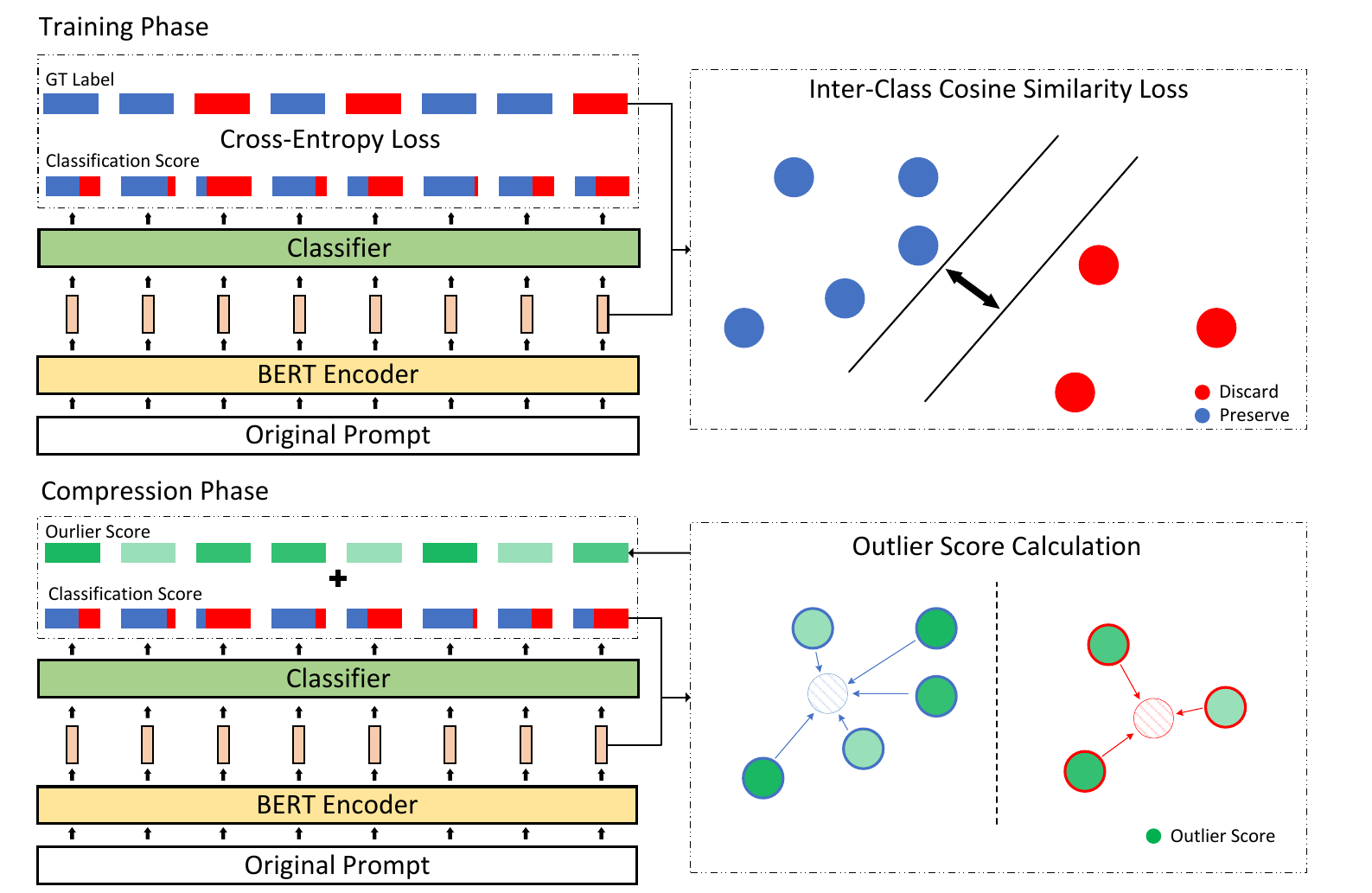}
  \caption{An overview of the proposed MOOSComp. \textbf{Training Phase (Top):} The compressor, containing a BERT-based encoder and a classifier in series, is trained for binary token classification with the cross-entropy loss and the inter-class cosine similarity loss to mitigate over-smoothing. \textbf{Compression Phase (Bottom):} Each token's classification score is combined with an outlier score to enhance the preservation of rare but important tokens.}
  \label{fig:overview}
  \vspace{-5pt}
\end{figure*}

Several long-context compression methods have been proposed and can be categorized as soft and hard prompt methods~\citep{li2024promptcompressionlargelanguage}. Soft prompt methods compress input prompts into shorter sequences of vectors~\citep{mu2023learning,chevalier2023adapting,ge2024incontext,wang-etal-2024-context-former}.
They often involve joint fine-tuning with the target model that will process the compressed inputs for specific tasks, which complicates the whole system. 
Conversely, hard prompt methods directly filter input prompts in a task-aware or task-agnostic manner. 
Task-aware methods~\citep{jiang-etal-2024-longllmlingua,jung2024discrete,CPC}, though effective, require intensive task-specific training or iterative inference. 
These requirements reduce task adaptability or increase compression cost.
In contrast, task-agnostic methods~\citep{li-etal-2023-compressing,jiang-etal-2023-llmlingua,pan-etal-2024-llmlingua} are more flexible and efficient as specific task knowledge is not required.

A notable task-agnostic hard prompt method is LLMLingua-2~\citep{pan-etal-2024-llmlingua}. It employs a BERT-based compressor~\citep{devlin2019bert,conneau2020unsupervised} trained on a text compression dataset to compress prompts through binary token classification, i.e., ``preserve'' or ``discard''. The compressed prompts are then fed into the target model for task execution. The advantages of this method are the minimal compression cost and direct applicability to various tasks. They are particularly desirable for resource-constrained scenarios.

Although this method has shown promise, it still face challenges regarding the token classification accuracy and generalization capability. Among them, the over-smoothing phenomenon of BERT-based models~\citep{dong2021attention,shi2022revisiting} stands out, where the token representations gradually converge to similar values as they pass through the layers. This can hinder the compressor's ability to distinguish tokens from different classes. To address this issue, we propose an anti-over-smoothing mechanism that adopts an inter-class cosine similarity loss term when training the compressor for better separating two classes.

We also note that since the compressor is trained on a specific dataset, strictly adhering to its output is suboptimal when predicting new tasks or data. Additionally, due to the task-agnostic nature of token selection, important tokens that are rare or deemed less important in the training data may be discarded during compression, even if they could be crucial for tasks at hand.
To preserve them, we propose an outlier detection mechanism that computes outlier scores per token to integrate with the classifier’s probability output. This ensures that tokens with high outlier scores, indicating their potential importance, are more likely to be preserved.

In summary, we propose MOOSComp to enhance the performance of a BERT-based long-context \textbf{Comp}ressor by \textbf{M}itigating the \textbf{O}ver-smoothing problem and incorporating \textbf{O}utlier \textbf{S}cores. An overview of MOOSComp is illustrated in Figure~\ref{fig:overview}. During the training of the compressor, the introduced inter-class cosine similarity loss eases token classification by penalizing excessive inter-class similarity in the output token representations. In the compression phase, the proposed outlier scores rescue rare but task-critical tokens to improve generalizability. 
Extensive experiments on four benchmarks including but not limited to long-context understanding and reasoning tasks show that MOOSComp consistently outperforms the SOTA task-agnostic hard prompt methods for either black-box API models or local models. Meanwhile, we achieve a speedup of 3.2x at a compression ratio of 5x on a V100 GPU and of 3.3x at a compression ratio of 4x on a smartphone.

\section{Related Work}

\noindent\textbf{Long-Context Compression.} Long-context compression is essential for the efficient use of LLMs, especially in resource-constrained environments, with mainly two types~\citep{li2024promptcompressionlargelanguage}.

\textit{1) Soft prompt methods} reduce the input size by representing the context with a sequence of continuous token representations. 
AutoCompressors~\citep{chevalier2023adapting} splits the input prompt into multiple segments and compresses them iteratively. 
ICAE~\citep{ge2024incontext} divides the training of compressor into pre-training and instruction fine-tuning, allowing the target model to be frozen and achieving better results. IC-Former~\citep{wang-etal-2024-context-former} employs a more lightweight compressor that incorporates only the cross-attention mechanism, largely reducing the compression cost.

\textit{2) Hard prompt methods}, in contrast, directly remove unnecessary input tokens. One class of these methods tailors compression to specific tasks. LongLLMLingua~\citep{jiang-etal-2024-longllmlingua} is designed for the task of question-relevant key information retrieval, inputting the long context along with the question to the compressor and calculating the perplexity of each token under different conditions to select tokens. PCRL~\citep{jung2024discrete} edits the prompts with a policy network trained on downstream tasks. CPC~\citep{CPC} scores the relevance of each sentence in a prompt to a specific question to determine token retention. 
Task-agnostic methods, on the contrary, do not rely on task-specific information. 
For example, Selective Context~\citep{li-etal-2023-compressing} calculates the self-information of lexical units and retains highly valued ones. LLMLingua~\citep{jiang-etal-2023-llmlingua} follows by proposing an iterative token-level compression algorithm to capture the inter-token dependencies. LLMLingua-2~\citep{pan-etal-2024-llmlingua} further reduces the compression cost by using a BERT-based compressor to perform token classification. Our approach improves LLMLingua-2 in both the training and compression phases, resulting in better compression performance and generalization capability.

\noindent\textbf{Over-Smoothing in BERT.} \citet{shi2022revisiting} point out the over-smoothing problem in BERT~\citep{devlin2019bert,liu2019roberta,Lan2020ALBERT}, where token representations become increasingly similar as the number of layers increases. They provide a theoretical explanation for its occurrence. 
To address this, \citet{guo2023contranorm} propose inserting contrastive normalization layers to prevent dimensional collapse of the representation matrix, while \citet{nguyen2023mitigating} propose adding the token representation difference between the first and current layers to the output of self-attention. \citet{pmlr-v235-zhou24g} modify the softmax function to achieve higher multi-modality and sparsity at the same time, thus mitigating over-smoothing. For the token-classification-based compression task, we analyze the similarity between token representations at each layer. Based on this analysis, we propose incorporating an inter-class cosine similarity loss during training to ease the token classification.

\noindent\textbf{Outlier Detection.} Outlier detection aims to identify data points, named outliers, that deviate significantly from the general data distribution~\citep{chandola2009anomaly}. Various detection methods have been proposed, generally in three types.
\textit{1) Distance-based methods}, of which a typical representative is K-Nearest Neighbors (KNN)~\citep{ramaswamy2000efficient,sun2022out} that defines the outlier score of a data point as the distance to its k-th nearest neighbor. 
\textit{2) Density-based methods}, such as LOF~\citep{breunig2000lof} and LoOP~\citep{kriegel2009loop}, classify points as outliers if they reside in lower-density regions.
\textit{3) Statistics-based methods}, such as Z-score~\citep{2019DetectingOI} and Mahalanobis distance~\citep{ghorbani2019mahalanobis}, measure how far a data point is from the mean of the dataset.
Defining outlier scores for token representations in a BERT-based compressor is a novel scenario. To minimize the additional time cost of compression, we adopt a simple yet effective method for scoring.

\section{Methodology}

\subsection{Preliminary: Prompt Compression Design in LLMLingua-2}\label{sec:preliminary}

LLMLingua-2~\citep{pan-etal-2024-llmlingua} is a task-agnostic approach, where a BERT-based compressor is used to perform token-level binary classification. To train the compressor, the authors construct a text compression dataset based on MeetingBank~\citep{hu-etal-2023-meetingbank} by using a specific instruction for GPT-4~\citep{achiam2023gpt} to compress the text.
Each token in the original texts is assigned a binary label (``preserve'' or ``discard'') depending on whether it appears in the compressed counterparts. The compressor, consisting of a pre-trained BERT model followed by a classification layer, is then trained on this dataset to perform binary classification on each token in the input prompt. The trained compressor can be applied to new data and tasks for prompt compression based on the predicted probability of each token being labeled as ``preserve''. Tokens with the highest probabilities are kept according to a pre-configured compression ratio. The advantages of LLMLingua-2 include a short compression time and strong applicability across different tasks. It strikes a valuable balance between compression quality and deployment efficiency.

\begin{figure}[t]
  \centering
  \includegraphics[width=0.9\columnwidth]{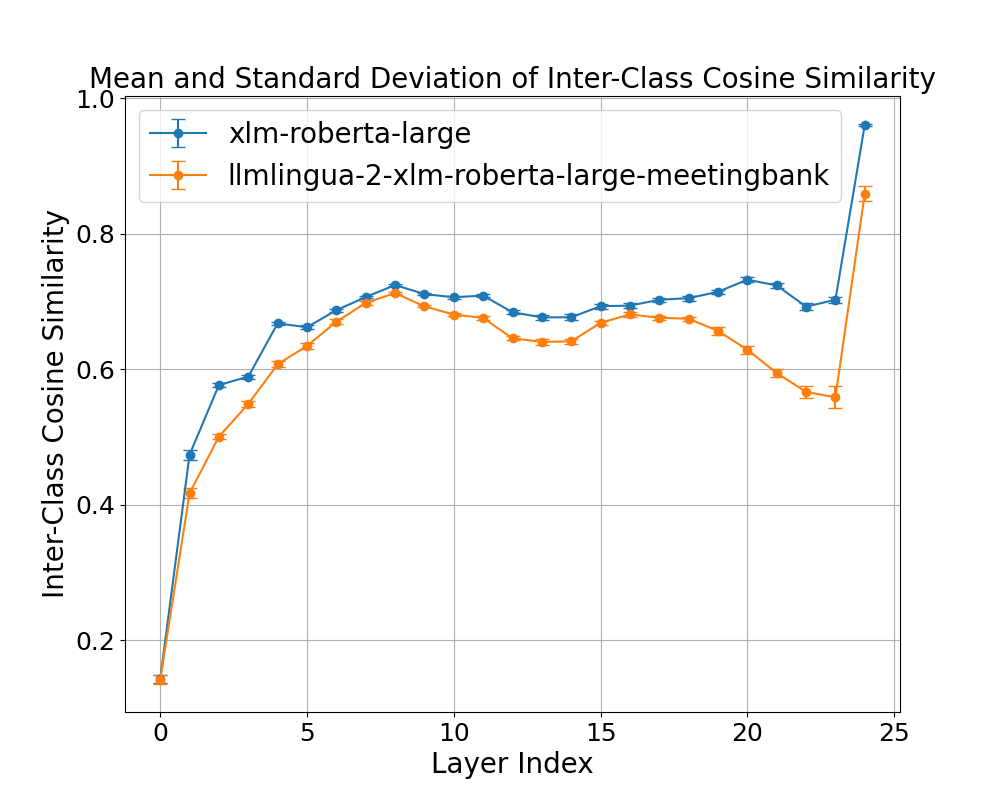}
  \caption{Inter-class cosine similarity of different layers in xlm-roberta-large and in llmlingua-2-xlm-roberta-large-meetingbank.}
  \label{fig:similarity}
  \vspace{-10pt}
\end{figure}

\subsection{Over-Smoothing in BERT-Based Compressor}\label{sec:oversmoothing}

The advantage of LLMLingua-2 in using a BERT-based compressor is that BERT has fewer parameters and faster inference speed than LLMs, while effectively leveraging bidirectional attention to capture inter-token relationships. However, BERT-based models, such as BERT~\citep{devlin2019bert}, RoBERTa~\citep{liu2019roberta} and ALBERT~\citep{Lan2020ALBERT}, suffer from over-smoothing~\citep{dong2021attention,shi2022revisiting}, where token representations become identical. 
This results in a loss of token distinctiveness, which make classification tasks, such as identifying which tokens to preserve or to discard, much more difficult.

To check whether over-smoothing exists in LLMLingua-2's BERT model, we examine the cosine similarity between the intermediate token representations for each layer. Since this model is for binary token classification, we mainly focus on the inter-class cosine similarity. Specifically, given an original prompt $\bm{x}=\{x_i\}_{i=1}^n$ with class labels $\bm{y}=\{y_i\}_{i=1}^n$, where $x_i$ is a token with label $y_i\in\{\text{``preserve''},\text{``discard''}\}$ and $n$ is the number of tokens, let $\bm{H^l}=\{\bm{h^l}_i\}_{i=1}^n$ denote the token representations produced by the $l$-th layer and the inter-class cosine similarity of layer $l$ is defined as

\noindent
\scalebox{0.9}{\parbox{1.11\columnwidth}{
\begin{equation}\label{eq:cossim}
  S^l = \frac{1}{|I_p| |I_d|} \sum_{i\in I_p,j\in I_d}\frac{\bm{h^l}_i \cdot \bm{h^l}_j}{\|\bm{h^l}_i\|_2 \|\bm{h^l}_j\|_2},
\end{equation}}}
where $I_p = \{ i \mid y_i=\text{``preserve''} \}$ and $I_d = \{ j \mid y_j=\text{``discard''} \}$ are the sets of indices.

We investigate whether the pre-trained BERT model xlm-roberta-large~\citep{conneau2020unsupervised} used in LLMLingua-2 and the fine-tuned model llmlingua-2-xlm-roberta-large-meetingbank~\citep{pan-etal-2024-llmlingua} trained by LLMLingua-2 exhibit the issue of over-smoothing. 
Using the data from the MeetingBank compression dataset mentioned in Section~\ref{sec:preliminary} as input, Figure~\ref{fig:similarity} shows the variation in the mean and standard deviation of the inter-class cosine similarities as the number of layers increases. Evidently, the original token representations of the two classes, corresponding to layer 0, are significantly different. However, as the model deepens, the similarity increases rapidly and is close to one in the last layer. From the results of llmlingua-2-xlm-roberta-large-meetingbank, we note that although the token classification training alleviates over-smoothing in the later layers, the inter-class similarity in the last layer remains high.

In summary, over-smoothing still exists in the fine-tuned compressor, which would face the challenge of making full use of the labeled data to maximize its token classification potential.

\subsection{Inter-Class Cosine Similarity}

Although there are several proposals to address over-smoothing~\citep{shi2022revisiting,guo2023contranorm,nguyen2023mitigating,pmlr-v235-zhou24g}, more targeted solutions can be devised for token classifiers in the task of prompt compression. 
Specifically, the objective of this task is clearly defined as token classification and enhancing the distinction between the representations of the two token classes by leveraging the labels could potentially improve classification performance. Furthermore, as analyzed in Section~\ref{sec:oversmoothing}, the token representations in the last layer exhibit much higher similarity compared with previous layers, and the classifier performs token classification based on these last-layer representations. Therefore, we propose to minimize the last-layer inter-class cosine similarity $S^L$ directly, where $L$ is the number of layers.

Specifically, let $f_{\bm{\varphi}}$ and $g_{\bm{\psi}}$ be the BERT-based encoder and the classifier on top of it with parameters $\bm{\varphi}$ and $\bm{\psi}$, respectively. The token representation produced by $f_{\bm{\varphi}}$ is $\bm{H^L}=\{\bm{h^L}_i\}_{i=1}^n = f_{\bm{\varphi}}(\bm{x})$. Then the inter-class cosine similarity loss is

\noindent
\scalebox{0.9}{\parbox{1.11\columnwidth}{
\begin{equation}\label{eq:cossimloss}
  \mathcal{L}_{\textbf{CS}}(\bm{\varphi}) \coloneqq S^L = \frac{1}{|I_p| |I_d|} \sum_{i\in I_p,j\in I_d}\frac{\bm{h^L}_i \cdot \bm{h^L}_j}{\|\bm{h^L}_i\|_2 \|\bm{h^L}_j\|_2}.
\end{equation}}}
For token representations in the ``preserve'' and ``discard'' classes, we compute the average cosine similarity between all pairs of tokens from these two classes. This additional loss helps to maintain sufficient inter-class separation and to improve the classification accuracy. It is simple and efficient, without adding inference-time overhead.
For those BERT-based models whose other layers exhibit the same issue, the proposed loss can be easily applied as well.

In addition to this newly introduced loss, we maintain the cross-entropy loss for the original token classification task, which is computed as 

\noindent
\scalebox{0.9}{\parbox{1.11\columnwidth}{
\begin{equation}\label{eq:celoss}
  \mathcal{L}_{\textbf{CE}}(\bm{\varphi},\bm{\psi}) = \frac{1}{n} \sum_{i=1}^n\ell(g_{\bm{\psi}}(\bm{h^L}_i), y_i),
\end{equation}}}
where $\ell(g_{\bm{\psi}}(\bm{h^L}_i), y_i)$ is the cross-entropy loss on one token $x_i$.
The final loss function combines the above two loss terms as follows:

\noindent
\scalebox{0.9}{\parbox{1.11\columnwidth}{
\begin{equation}\label{eq:finalloss}
  \mathcal{L}(\bm{\varphi},\bm{\psi}) = \mathcal{L}_{\textbf{CE}}(\bm{\varphi},\bm{\psi}) + \beta \mathcal{L}_{\textbf{CS}}(\bm{\varphi}),
\end{equation}}}
where $\beta$ is a hyperparameter that balances the two terms. This not only ensures the model excels in classification, but also prevents over-smoothing.

\subsection{Outlier Scores}

Despite its large training corpus, the aforementioned prompt compressor cannot yet cover all task types. Therefore, relying solely on its classification is likely to be suboptimal for new tasks and datasets. Additionally, for task-agnostic methods, since compression does not consider task-specific information, there is a high risk of removing task-relevant tokens that could appear infrequently or are considered unimportant in the general corpus. This has been seen in certain tasks, such as long-document question answering or summarization.
To retain these tokens, we introduce a mechanism to assign outlier scores per token and preserve highly-scored ones. This idea aligns with previous methods~\citep{li-etal-2023-compressing,jiang-etal-2023-llmlingua} that leverage LLMs to retain high-perplexity tokens.

When designing the outlier detection method, we focus on two key factors: 1) minimal introduction of hyperparameters, and 2) minimal computational overhead. We aim to avoid methods such as KNN~\citep{ramaswamy2000efficient,sun2022out} or LOF~\citep{breunig2000lof}, which introduce additional hyperparameters (e.g., the number of neighbors in KNN) and require careful tuning. 
We would also avoid significant computational overhead, such as matrix factorization or matrix inversion~\citep{ghorbani2019mahalanobis}, which increases compression cost.

In the following, we introduce the calculation method for the outlier scores. Since we use inter-class cosine similarity loss during training to increase the inter-class distance, 
it would be inappropriate to estimate a single distribution over all the token representations and compute outlier scores based on that. 
Therefore, we first divide the tokens into two categories based on the classifier's output, and then compute outlier scores separately within each category. To be more specific, let
$p_i^{\text{preserve}} = p(y=\text{``preserve''} \mid x_i,\ g_{\bm{\psi}} \circ f_{\bm{\varphi}})$ denote the probability that the compressor $g_{\bm{\psi}} \circ f_{\bm{\varphi}}$ classifies the token $x_i$ as ``preserve''.
The tokens can then be divided into two categories, with the corresponding sets of indices denoted as $I'_p = \{ i \mid p_i^{\text{preserve}} \geq 0.5 \}$ and $I'_d = \{ j \mid p_j^{\text{preserve}} < 0.5 \}$. 
We propose to compute outlier scores based on Z-score~\citep{2019DetectingOI}. For tokens with indices belonging to $I'_p$, the Z-score is calculated as

\noindent
\scalebox{0.9}{\parbox{1.11\columnwidth}{
\begin{equation}\label{eq:zscore}
  \bm{z}_i = \frac{\bm{h^L}_i-\bm{\mu_p}}{\bm{\sigma_p}},
\end{equation}}}
where $\bm{\mu_p}=\frac{1}{|I'_p|}\sum_{i\in I'_p}\bm{h^L}_i$ is the mean of such representations and $\bm{\sigma_p}=\sqrt{\frac{1}{|I'_p|}\sum_{i\in I'_p}(\bm{h^L}_i-\bm{\mu_p})^2}$ is the standard deviation. Similarly, for tokens whose indices belong to $I'_d$, the Z-score is calculated as $\bm{z}_j = \frac{\bm{h^L}_j-\bm{\mu_d}}{\bm{\sigma_d}}$, where $\bm{\mu_d}=\frac{1}{|I'_d|}\sum_{j\in I'_d}\bm{h^L}_j$ and $\bm{\sigma_d}=\sqrt{\frac{1}{|I'_d|}\sum_{j\in I'_d}(\bm{h^L}_j-\bm{\mu_d})^2}$.

Since $\bm{z}_i$ is a high-dimensional vector, to account for the standard deviations in all dimensions, we use the Euclidean norm $s_i=\|\bm{z}_i\|_2$ as the score for token $x_i$. To make the outlier scores compatible with the classifier's probability output, we normalize them to the range $[0, 1]$~\citep{kriegel2011interpreting}:

\noindent
\scalebox{0.9}{\parbox{1.11\columnwidth}{
\begin{equation}\label{eq:outlierscore}
  s_i^{\text{norm}} = \frac{s_i - s_p^{\text{min}}}{s_p^{\text{max}} - s_p^{\text{min}}},
\end{equation}}}
where $s_p^{\text{min}}$ and $s_p^{\text{max}}$ are the minimum and maximum values of $\{s_i\}_{i\in I'_p}$, respectively. Similarly, for $x_j$ whose $j \in I'_d$, the normalized outlier score is calculated as $s_j^{\text{norm}} = \frac{s_j - s_d^{\text{min}}}{s_d^{\text{max}} - s_d^{\text{min}}}$, where $s_j=\|\bm{z}_j\|_2$, and $s_d^{\text{min}}$ and $s_d^{\text{max}}$ are the minimum and maximum values of $\{s_j\}_{j\in I'_d}$, respectively.

Finally, the outlier scores are integrated with the classifier’s output to form the compression metric:

\noindent
\scalebox{0.9}{\parbox{1.11\columnwidth}{
\begin{equation}\label{eq:compressionmetric}
  m_k = \alpha p_k^{\text{preserve}} + (1-\alpha) s_k^{\text{norm}},
\end{equation}}}
where $\alpha$ is a hyperparameter that balances $p_k^{\text{preserve}}$ and $s_k^{\text{norm}}$, and $k\in I'_p \cup I'_d$.
This weighted combination allows the model to prioritize rare but important tokens while still honoring the classifier.

During the compression phase, according to this metric, the top $\hat{n}$ tokens with the highest values are retained, where $\hat{n}$ is determined based on a predefined compression ratio.

\section{Experiments}

In this section, we introduce the experimental setup and present the main results, including the applicability of the compressor to various data, tasks, and target models, as well as the inference speedup achieved through prompt compression. We also include ablation studies to validate the effectiveness of the newly introduced modules.

\subsection{Implementation Details}

\noindent\textbf{Compressor and Target Models.} Following LLMLingua-2~\citep{pan-etal-2024-llmlingua}, we use xlm-roberta-large~\citep{conneau2020unsupervised} as the compressor's feature encoder, connected to a linear classification layer. To verify whether the compressor applies to different target models, we examine both black-box API models, GPT-3.5-Turbo-0613 and GPT-4o-mini~\citep{achiam2023gpt}, and local models of various sizes, including Qwen2.5-7B-Instruct and Qwen2.5-3B-Instruct~\citep{yang2024qwen2}.

\begin{figure}[t]
  \centering
  \includegraphics[width=0.9\columnwidth]{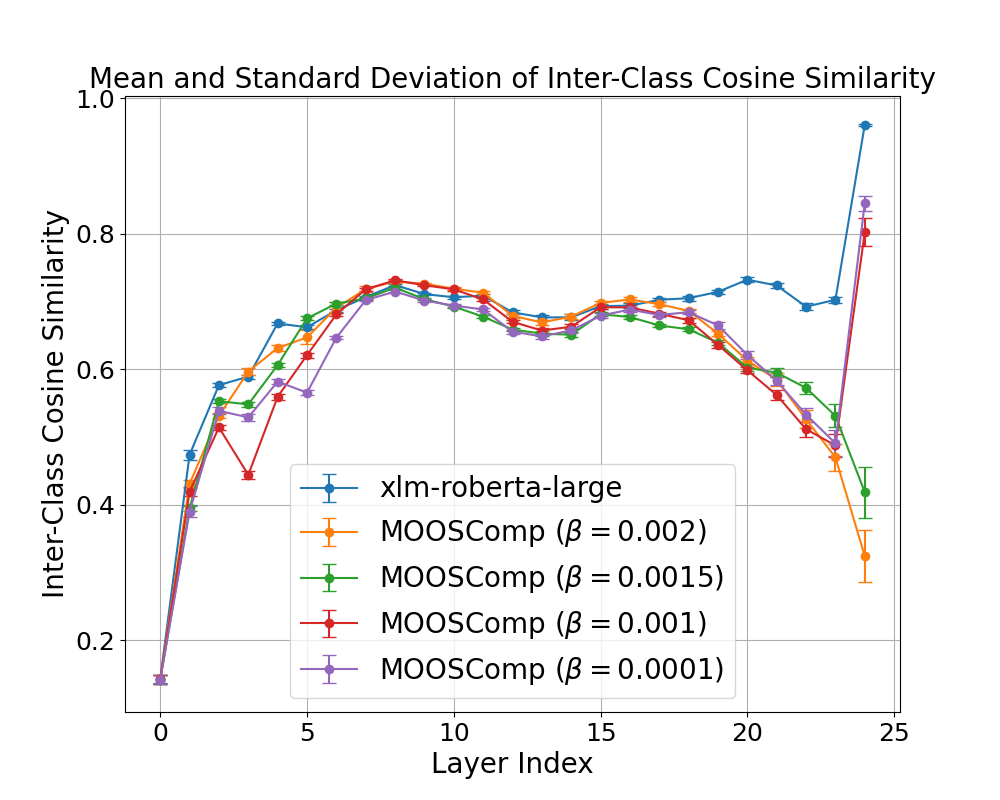}
  \caption{Inter-class cosine similarity of different layers in xlm-roberta-large and in MOOSComp with varying values of $\beta$.}
  \label{fig:similarity_beta}
\end{figure}

\begin{table}[tp]
    \centering
    \addtolength{\tabcolsep}{-0.4em}
    \makebox[\columnwidth]{\resizebox{1.02\columnwidth}{!}{
    \begin{tabular}{lccccccc}
    \hline
    \multirow{3}{*}{\textbf{Methods}} & \multicolumn{5}{c}{\textbf{Summary}} & \multicolumn{2}{c}{\textbf{Length}} \\
    \cmidrule(lr){2-6} \cmidrule(lr){7-8} 
    & BLEU & Rouge1 & Rouge2 & RougeL & BERTScore & Tokens & $1/\tau$ \\
    \hline
    Selective Context & 8.03 & 41.92 & 15.79 & 27.95 & 88.02 & 1,049 & 3x \\
    LLMLingua & 7.80 & 42.13 & 14.89 & 28.30 & 87.97 & 1,186 & 3x \\
    LLMLingua-2-small & 18.38 & 53.15 & 25.56 & 37.00 & 89.99 & 895 & 3x \\
    LLMLingua-2 & 19.89 & 54.50 & 26.91 & 38.34 & 90.27 & 982 & 3x \\
    \textbf{MOOSComp} & \textbf{20.50} & \textbf{54.72} & \textbf{27.98} & \textbf{39.32} & \textbf{90.36} & 991 & 3x \\
    \hline
    Original & 27.52 & 58.50 & 35.12 & 44.98 & 91.40 & 3,003 & 1x \\
    \hline
    \end{tabular}
    }}
    \caption{Results of MOOSComp and other methods on MeetingBank using GPT-3.5-Turbo.}
    \label{tab:meetingbank_evaluation}
    \vspace{-10pt}
\end{table}

\noindent\textbf{Datasets and Tasks.} Our compressor is trained on the MeetingBank compression dataset~\citep{pan-etal-2024-llmlingua} mentioned in Section~\ref{sec:preliminary}. 
After training, we first evaluate its impact on the summarization task using the in-domain MeetingBank test data~\citep{hu-etal-2023-meetingbank}. Then we assess the effect on various out-of-domain tasks, including long-context understanding on LongBench~\citep{bai-etal-2024-longbench}, math reasoning on GSM8K~\citep{cobbe2021gsm8k}, and language reasoning on BBH~\citep{srivastava2023beyond}.
These datasets cover different domains, tasks, and language styles that the compressor has not been exposed to during training. By including such evaluations, we aim to assess the generalization ability of our compression method, which is critical for real-world applications where the model is often deployed in unseen domains.
Appendix~\ref{sec:dataset} has more details. All the tasks use the same evaluation metrics as LLMLingua-2.

\noindent\textbf{Baselines.} Our comparison is mainly among task-agnostic hard prompt methods, including Selective Context~\citep{li-etal-2023-compressing}, LLMLingua~\citep{jiang-etal-2023-llmlingua} and LLMLingua-2. The first two compress with LLaMA-2-7B~\citep{touvron2023llama}, while the last uses multilingual-BERT~\citep{devlin2019bert} and xlm-roberta-large~\citep{conneau2020unsupervised}.

\begin{table}[tp]
    \centering
    \addtolength{\tabcolsep}{-0.4em}
    \makebox[\columnwidth]{\resizebox{1.02\columnwidth}{!}{
    \begin{threeparttable}
    \begin{tabular}{lccccccccc}
    \hline
    \multirow{3}{*}{\textbf{Methods}} & \multicolumn{9}{c}{\textbf{LongBench}}  \\
    \cmidrule(lr){2-10} 
                     & SingleDoc & MultiDoc & Summ. & FewShot & Synth. & Code & AVG & Tokens & $1/\tau$ \\
    \hline
    \hline
                     & \multicolumn{9}{c}{\textit{2,000-token constraint}} \\
    \hline
    Selective Context & 16.2 & 34.8 & 24.4 & 15.7 & 8.4 & 49.2 & 24.8 & 1,925 & 5x \\
    LLMLingua & 22.4 & 32.1 & 24.5 & 61.2 & 10.4 & 56.8 & 34.6 & 1,950 & 5x \\
    LLMLingua-2-small & 29.5 & 32.0 & 24.5 & 64.8 & 22.3 & 56.2 & 38.2 & 1,891 & 5x  \\
    LLMLingua-2 & 29.8 & 33.1 & 25.3 & 66.4 & 21.3 & \textbf{58.9} & 39.1 & 1,954 & 5x  \\
    \textbf{MOOSComp} & \textbf{35.3} & \textbf{35.5} & \textbf{26.3} & \textbf{66.6} & \textbf{37.5} & 53.8 & \textbf{42.5} & 1,975 & 5x  \\
    \hline
    \hline
                     & \multicolumn{9}{c}{\textit{3,000-token constraint}} \\
    \hline
    Selective Context & 23.3 & \textbf{39.2} & 25.0 & 23.8 & 27.5 & 53.1 & 32.0 & 3,328 & 3x \\
    LLMLingua & 31.8 & 37.5 & 26.2 & 67.2 & 8.3 & 53.2 & 37.4 & 3,421 & 3x \\
    LLMLingua-2-small & 35.5 & 38.1 & 26.2 & 67.5 & 23.9 & 60.0 & 41.9 & 3,278 & 3x \\
    LLMLingua-2 & 35.5 & 38.7 & 26.3 & \textbf{69.6} & 21.4 & \textbf{62.8} & 42.4 & 3,392 & 3x  \\
    \textbf{MOOSComp} & \textbf{37.5} & 38.4 & \textbf{27.2} & 67.0 & \textbf{41.5} & 61.0 & \textbf{45.4} & 2,934 & 4x  \\
    \hline
    \hline
    Original Prompt & 39.7 & 38.7 & 26.5 & 67.0 & 37.8 & 54.2 & 44.0 & 10,295 & 1x \\
    \hline
    Zero-Shot & 15.6 & 31.3 & 15.6 & 40.7 & 1.6 & 36.2 & 23.5 & 214 & 48x \\
    \hline
    \end{tabular}
    \end{threeparttable}
    }}
    \caption{Out-of-domain evaluation under different compression ratios on LongBench using GPT-3.5-Turbo.}
    \label{tab:long_context}
\end{table}

\begin{table}[tp]
    \centering
    \resizebox{\columnwidth}{!}{
    \begin{tabular}{lcccccccccccc}
    \hline
    \multirow{3}{*}{\textbf{Methods}} & \multicolumn{3}{c}{\textbf{GSM8K}} & \multicolumn{3}{c}{\textbf{BBH}} \\
    \cmidrule(lr){2-4} \cmidrule(lr){5-7}
    & EM & Tokens & $1/\tau$ & EM & Tokens & $1/\tau$ \\
    \hline
    \hline
    & \multicolumn{6}{c}{\textit{half-shot constraint}} \\
    \hline
    Selective Context & 80.74 & 215 & 11x & 54.02 & 188 & 4x \\
    LLMLingua & 83.24 & 172 & 14x & 61.21 & 193 & 4x \\
    LLMLingua-2-small & 83.24 & 167 & 14x & 58.98 & 171 & 4x \\
    LLMLingua-2 & 82.94 & 179 & 13x & 61.25 & 153 & 5x \\
    \textbf{MOOSComp} & \textbf{84.46} & 176 & 13x & \textbf{63.24} & 156 & 5x \\
    \hline
    \hline
    Full-Shot & 86.13 & 2,359 & 1x & 69.04 & 758 & 1x \\
    \hline
    \end{tabular}
    }
    \caption{Out-of-domain evaluation of different methods on GSM8K and BBH using GPT-4o-mini.}
    \label{tab:reasoning_evaluation}
    \vspace{-12pt}
\end{table}

\subsection{Main Results}\label{subsec:results}

\noindent\textbf{Impact of Inter-Class Cosine Similarity Loss.} We first validate the effectiveness of our proposed anti-over-smoothing mechanism.
We divide the MeetingBank compression dataset into 80\% for training the compressor and 20\% for validation. We adopt different values of $\beta$ in Eq.~\eqref{eq:finalloss}, yielding different compressors. Additional training details can be found in Appendix~\ref{sec:training}.
After training, we calculate the inter-class cosine similarity per layer for both our trained models and the pre-trained BERT model xlm-roberta-large.
The results are depicted in Figure~\ref{fig:similarity_beta}. When $\beta$ is larger, the similarity in the last layer decreases drastically. This shows that adjusting $\beta$ effectively alleviates the over-smoothing issue.
Based on the models' validation accuracies, we select the one trained with $\beta = 0.001$ for further evaluations in the remaining experiments.

\noindent\textbf{In-Domain Evaluation with Black-Box API Models.} On the MeetingBank test set, we use the compressor to reduce the token count to $\tau=1/3$ of the original and then feed the compressed input into GPT-3.5-Turbo for summarization. We determine the optimal $\alpha$ in Eq.~\eqref{eq:compressionmetric} using 20\% of the test data and then apply it to the entire dataset.
Interestingly, for this task, the optimal value of $\alpha$ is 0.5, that is, the probability output of the classifier and the outlier scores contribute equally. The evaluation results 
are presented in Table~\ref{tab:meetingbank_evaluation}. MOOSComp outperforms all baselines including LLMLingua-2, indicating the effectiveness of the newly introduced loss function and outlier detection mechanism.

\begin{table}[tp]
    \centering
    \resizebox{\columnwidth}{!}{
    \begin{tabular}{lccccc}
    \hline
    \multirow{3}{*}{\textbf{Methods}} & \multicolumn{5}{c}{\textbf{LongBench}}  \\
    \cmidrule(lr){2-6} 
                                      & SingleDoc & MultiDoc & Summ. & Tokens & $1/\tau$ \\
    \hline
    \hline
    Qwen2.5-7B-Instruct    & \multicolumn{5}{c}{\textit{2,000-token constraint}} \\
    \hline
    Selective Context & 18.5 & 25.8 & 23.7 & 2,041 & 6x \\
    LLMLingua & 26.5 & 23.9 & 24.5 & 2,124 & 5x \\
    LLMLingua-2-small & 28.6 & 34.3 & 26.0 & 2,068 & 6x  \\
    LLMLingua-2 & 31.5 & 32.9 & 26.5 & 2,060 & 6x  \\
    \textbf{MOOSComp} & \textbf{34.9} & \textbf{37.0} & \textbf{26.8} & 2,112 & 6x \\
    \hline
    \hline
    Qwen2.5-7B-Instruct    & \multicolumn{5}{c}{\textit{3,000-token constraint}} \\
    \hline
    Selective Context & 20.8 & 26.7 & 25.0 & 3,302 & 4x \\
    LLMLingua & 29.2 & 25.9 & 25.4 & 2,822 & 4x \\
    LLMLingua-2-small & 32.7 & 37.2 & 26.6 & 2,916 & 4x \\
    LLMLingua-2 & 34.6 & 36.2 & 27.1 & 2,951 & 4x  \\
    \textbf{MOOSComp} & \textbf{36.5} & \textbf{40.6} & \textbf{27.4} & 2,977 & 4x  \\
    \hline
    \hline
    Original Prompt & 42.0 & 45.0 & 28.2 & 11,668 & 1x \\
    \hline
    \hline
    Qwen2.5-3B-Instruct    & \multicolumn{5}{c}{\textit{2,000-token constraint}} \\
    \hline
    Selective Context & 15.4 & 20.6 & 22.2 & 2,041 & 6x \\
    LLMLingua & 22.8 & 16.0 & 23.3 & 2,124 & 5x \\
    LLMLingua-2-small & 23.5 & 24.8 & 25.0 & 2,068 & 6x  \\
    LLMLingua-2 & 25.4 & 26.7 & 25.0 & 2,060 & 6x  \\
    \textbf{MOOSComp} & \textbf{27.6} & \textbf{28.2} & \textbf{25.2} & 2,112 & 6x \\
    \hline
    \hline
    Qwen2.5-3B-Instruct    & \multicolumn{5}{c}{\textit{3,000-token constraint}} \\
    \hline
    Selective Context & 18.0 & 22.6 & 23.7 & 3,302 & 4x \\
    LLMLingua & 25.5 & 17.9 & 24.4 & 2,822 & 4x \\
    LLMLingua-2-small & 27.6 & \textbf{28.9} & 25.6 & 2,916 & 4x \\
    LLMLingua-2 & 27.1 & 28.5 & 25.9 & 2,951 & 4x  \\
    \textbf{MOOSComp} & \textbf{31.4} & 28.7 & \textbf{26.0} & 2,977 & 4x  \\
    \hline
    \hline
    Original Prompt & 36.2 & 35.0 & 27.4 & 11,668 & 1x \\
    \hline
    \end{tabular}
    }
    \caption{Out-of-domain evaluation under different compression ratios on LongBench using local models.}
    \label{tab:longbench_local}
    \vspace{-15pt}
\end{table}

\noindent\textbf{Out-of-Domain Evaluation with Black-Box API Models.} To validate the generalizability of our method on out-of-domain data and different target models, we evaluate it for long-context understanding tasks on LongBench with GPT-3.5-Turbo\footnote{For the code completion sub-task on LongBench, we use GPT-4o-mini as the target model, as we could only achieve an accuracy of 49.3 with the original prompt input to GPT-3.5-Turbo, which is much lower than the result in Table~\ref{tab:long_context}.}, as well as for reasoning tasks on GSM8K and BBH with GPT-4o-mini. For each task, we still use 20\% of the data to select the optimal $\alpha$.
We note that on most tasks, setting $\alpha$ to 0.8 and 0.7 yields the best performance.
The results on LongBench and on GSM8K/BBH are reported in Tables~\ref{tab:long_context} and \ref{tab:reasoning_evaluation}, respectively\footnote{Except for our results, those of the other methods on LongBench are from the LLMLingua-2 work.}. 
Our method significantly outperforms the others on most tasks. Inclusion of outlier scores makes it more likely to keep rare but important tokens. For instance, on the multi-document QA task, although our compressor does not know the question during prompt compression, the answer often contains tokens that are relatively rare in the general corpora and, therefore, they are more likely to be retained. Specific examples can be found in Appendix~\ref{sec:compressed_texts}. However, the proposed outlier detection mechanism does not benefit the code completion task. We suspect that this is due to the unique inherently structural nature of code texts. Retaining more tokens that encapsulate the code logic could be essential to achieve better results.

\begin{table}[tp]
    \centering
    \resizebox{\columnwidth}{!}{
    \begin{tabular}{lcccc}
    \hline
    $1/\tau$    &   2x   &   3x    &    4x     &     5x    \\  
    \hline
    \hline
    Qwen2.5-7B-Instruct  & \multicolumn{4}{c}{\textit{end-to-end latency (s)}} \\
    \hline
    Without Compression  & \multicolumn{4}{c}{8.14} \\
    With MOOSComp        &  4.62  &  3.49   &  2.94  &  2.54 \\
    Speedup              &  1.8x  &  2.3x   &  2.8x  &  3.2x \\
    \hline
    \hline
                         & \multicolumn{4}{c}{\textit{compression cost (s)}} \\
    \hline
    Selective Context    & 5.52 & 5.53 & 5.53 & 5.53 \\
    LLMLingua            & 4.14 & 3.86 & 3.71 & 3.63 \\
    LLMLingua-2          & 0.40 & 0.40 & 0.40 & 0.40 \\
    MOOSComp             & 0.41 & 0.41 & 0.41 & 0.40 \\
    \hline
    \end{tabular}
    }
    \caption{Latency evaluation under different compression ratios on the single-document QA task of LongBench.}
    \label{tab:latency_gpu}
\end{table}

\begin{table}[tp]
    \centering
    \resizebox{\columnwidth}{!}{
    \begin{tabular}{lccc|ccc}
    \hline
    \# Original Tokens       &   \multicolumn{3}{c}{2.7K}  &   \multicolumn{3}{c}{6.7K}    \\
    \hline
    $1/\tau$       &   1x  &   3x    &    4x  &   1x  &   2x    &    4x\\
    \hline
    Compression (s)  &  -  &   1.89   &  1.89   &  -  &   4.42   &  4.42  \\
    Prefill (s)      &  4.16  &   0.65   &  0.55  &  22.17  &   5.36   &  1.27  \\
    Decoding (s)     &  3.62  &   3.48   &  3.10  &  7.54   &   3.91   &  3.24  \\
    Total (s)        &  7.78 &    6.02   &  5.54  &  29.71  &   13.69  &  8.93  \\
    Speedup          &  1x    &   1.3x   &  1.4x  &  1x     &   2.2x   &  3.3x \\
    \hline
    \end{tabular}
    }
    \caption{Latency evaluation of the compressor and Qwen2.5-3B-Instruct on a smartphone.}
    \label{tab:latency_npu}
    \vspace{-15pt}
\end{table}

\noindent\textbf{Out-of-Domain Evaluation with Local Models.} As the technology for miniaturizing LLMs continues to mature, there is an increasing demand for deploying relatively smaller LLMs in resource-constrained environments, such as PCs and smartphones. Prompt compression can effectively reduce model inference time and memory usage, while our method is well-suited for resource-limited settings. 
Here, we validate the effectiveness of MOOSComp targeting Qwen2.5-7B-Instruct and Qwen2.5-3B-Instruct. Table~\ref{tab:longbench_local} shows the results with our method consistently excelling. Although the training data are generated with GPT-4, the resulting compressor applies to other families of models as well.

\noindent\textbf{Latency Evaluation on NVIDIA GPUs.} To evaluate the inference time savings of Qwen2.5-7B-Instruct after prompt compression in low-resource environments, we test latency on a V100-32G GPU for the single-document QA task on LongBench. We choose examples with fewer than 6.5K tokens to avoid hitting the GPU memory limits.
Table~\ref{tab:latency_gpu} presents the end-to-end latency using MOOSComp, along with the compression costs of various methods. 
Our approach achieves a speedup of 1.8x-3.2x. Furthermore, due to the efficiency of our proposed calculation of outlier scores, our method shows better accuracy than LLMLingua-2 while incurring almost no additional compression cost.

\begin{table}[t]
    \centering
    \resizebox{\columnwidth}{!}{
    \begin{tabular}{lcccc}
    \hline
    \multirow{3}{*}{\textbf{Methods}} & \multicolumn{4}{c}{\textbf{LongBench}}  \\
    \cmidrule(lr){2-5} 
                                      & \multicolumn{2}{c}{SingleDoc} & \multicolumn{2}{c}{MultiDoc} \\
    \hline
    \hline
    \textit{token constraint}    & \textit{2K} & \textit{3K} & \textit{2K} & \textit{3K}\\
    \hline
    LLMLingua-2 & 31.5 & 34.6 & 32.9 & 36.2  \\
    MOOSComp w/o $\mathcal{L}_{\textbf{CS}}$ & 32.3 & 36.5 & 34.7 & 38.2  \\
    MOOSComp w/o $s^{\text{norm}}$ & 32.6 & 35.8 & 35.7 & 39.3  \\
    \hline
    \hline
    MOOSComp w/ $\mathcal{L}_{\textbf{CS}}$ variant & 32.6 & 34.4 & 34.9 & 38.2  \\
    MOOSComp w/ Gate & 32.9 & 35.8 & 35.5 & 38.2  \\
    \hline
    \hline
    MOOSComp w/ $s^{\text{norm}}$ variant & 33.0 & \textbf{36.9} & 35.3 & 38.9  \\
    MOOSComp w/ KNN ($k=10$) & 33.6 & 36.2 & 35.5 & 39.3  \\
    MOOSComp w/ KNN ($k=20$) & 33.5 & 35.9 & 36.5 & 39.0  \\
    MOOSComp w/ KNN ($k=40$) & 33.5 & 35.4 & 35.2 & 39.6  \\
    \hline
    \hline
    \textbf{MOOSComp} & \textbf{34.9} & 36.5 & \textbf{37.0} & \textbf{40.6}  \\
    \hline
    \end{tabular}
    }
    \caption{Ablation study on LongBench using Qwen2.5-7B-Instruct.}
    \label{tab:ablation_contribution}
    \vspace{-15pt}
\end{table}

\noindent\textbf{Latency Evaluation on Mobile Devices.} To demonstrate the acceleration effects of our method on resource-constrained edge devices, we experiment with Qwen2.5-3B-Instruct deployed on the Neural Processing Unit (NPU, an embedded AI accelerator) of a smartphone. The deployment procedure involves the steps generally adopted by chip vendors, including model preparation, conversion, Post-Training Quantization (PTQ), compilation, and on-device inference. Please refer to Appendix~\ref{sec:llm_deployment} for more details. 
The BERT-based compressor is deployed on the smartphone GPU due to the flexibility and ease of use~\citep{li2024transformer}. We utilize mixed precision and dynamic shape model inference.
Appendix~\ref{sec:bert_deployment} has more details on this. To check the acceleration effects, we use LongBench summarization examples with 2.7K and 6.7K tokens, and compress them at ratios 2x, 3x, and 4x, respectively. 
Table~\ref{tab:latency_npu} shows the results of on-device inference. Thanks to the efficiency of MOOSComp, even with the additional compression costs, an overall speedup of 1.3x-3.3x can still be achieved. Acceleration is particularly pronounced during the prefill phase, reaching up to 3.9x, significantly reducing the time-to-first-token.

\subsection{Ablation Studies}

\noindent\textbf{Contribution of the Two Mechanisms.} To demonstrate the individual contributions of the anti-over-smoothing and outlier detection mechanisms, we compare two variants of our method: 
1) training the compressor without incorporating the inter-class cosine similarity loss (\textit{MOOSComp w/o $\mathcal{L}_{\textbf{CS}}$}); 
and 2) not using outlier scores during the compression phase (\textit{MOOSComp w/o $s^{\text{norm}}$}). 
We conduct experiments on LongBench with Qwen2.5-7B-Instruct, and the results are presented in Table~\ref{tab:ablation_contribution}. 
Compared with LLMLingua-2, the inclusion of either mechanism alone improves task performance. Furthermore, our proposed approach of incorporating both mechanisms simultaneously achieves the optimal results.

\noindent\textbf{Design of the Two Mechanisms.} We also compare the effects of different design strategies for the anti-over-smoothing and outlier detection mechanisms. For the former one, we compare \textit{MOOSComp w/o $s^{\text{norm}}$} with the following two methods: 1) a variant of our method that calculates the average cosine similarity between all pairs of token representations in the last layer (\textit{MOOSComp w/ $\mathcal{L}_{\textbf{CS}}$ variant}); and 2) the vertical gate fusion proposed by~\citet{shi2022revisiting}, which adaptively weighs and sums the representations of each token across all layers to produce the final output representation (\textit{MOOSComp w/ Gate}). For the latter one, we also compare two approaches: 1) a variant of our method that calculates outlier scores without dividing tokens into two categories (\textit{MOOSComp w/ $s^{\text{norm}}$ variant}); and 2) defining the outlier score of a token as the cosine distance to its k-th nearest neighbor (\textit{MOOSComp w/ KNN}). The results in Table~\ref{tab:ablation_contribution} indicate that reducing cosine similarity between all token pairs improves performance, though not as pronounced as the reduction in inter-class cosine similarity. The vertical gate fusion performs well. However, it requires the storage of all intermediate features during inference, which leads to reduced inference speed and increased memory usage, thereby increasing the compression cost. In contrast, our method introduces no additional inference-time cost, making it better aligned with the practical goal of building an efficient prompt compression module. 
For calculating outlier scores, performing separate computations for the two categories is more effective. Regarding KNN, we test various values of k, but the performance does not surpass that of MOOSComp. In summary, mitigating the over-smoothing issue and employing the outlier detection mechanism are both beneficial for improving accuracy, and our method is overall superior to other variants and approaches.

\section{Conclusion}\label{sec:conclusion}

In this paper, we have presented MOOSComp, a task-agnostic hard prompt method for efficient long-context compression that addresses critical challenges in token classification. By introducing the inter-class cosine similarity loss and the outlier detection mechanism, MOOSComp addresses the issue of over-smoothing in BERT-based models and the misclassification of rare yet important tokens, respectively. Extensive experiments over four benchmarks and four target models demonstrate that after resolving these issues, our method achieves superior prompt compression effectiveness and generalizability compared with existing methods. Our method also delivers significant acceleration in resource-constrained environments.

\bibliography{custom}

\appendix

\section{Dataset Details}\label{sec:dataset}

\noindent\textbf{MeetingBank.} MeetingBank~\citep{hu-etal-2023-meetingbank} is a benchmark dataset for meeting summarization, introduced to address the lack of annotated meeting corpora for developing summarization technology. We utilize the MeetingBank test set provided by LLMLingua-2~\citep{pan-etal-2024-llmlingua}\footnote{\url{https://huggingface.co/datasets/microsoft/MeetingBank-QA-Summary}} to evaluate the effectiveness of prompt compression on the meeting summarization task. It contains 862 meeting transcripts with summaries generated by GPT-4.

\noindent\textbf{LongBench.} LongBench~\citep{bai-etal-2024-longbench} is a benchmark dataset designed to evaluate the long-context understanding capabilities of large language models. For our experiments, we use 16 English tasks across six major categories, including single-document QA, multi-document QA, summarization, few-shot learning, synthetic tasks and code completion.

\noindent\textbf{GSM8K.} GSM8K~\citep{cobbe2021gsm8k} is a widely used benchmark for math reasoning and problem-solving, particularly useful for testing multi-step reasoning. 
In our experiments, we employ the GSM8K test set containing 1,319 problems for evaluation. 
Following LLMLingua~\citep{jiang-etal-2023-llmlingua}, we adopt the few-shot chain-of-thought prompt~\citep{fu2022complexity}\footnote{\url{https://github.com/FranxYao/chain-of-thought-hub}} as the original prompt.

\noindent\textbf{BBH.} BBH~\citep{suzgun-etal-2023-challenging} serves as a benchmark to assess the language and symbolic reasoning capabilities of large language models. It has a suite of 23 challenging tasks with 6,511 problems. We employ the 3-shot chain-of-thought prompt\footnote{\url{https://github.com/suzgunmirac/BIG-Bench-Hard}} provided by~\citet{suzgun-etal-2023-challenging} as the original prompt.

\section{Training Details}\label{sec:training}
We adopt xlm-roberta-large~\citep{conneau2020unsupervised} as the pre-trained feature encoder of our compressor. The number of parameters of this pre-trained model is 559M. Our compressor is trained on the MeetingBank compression dataset~\citep{pan-etal-2024-llmlingua} using PyTorch 2.1.2 on an A100-80G GPU. We divide the dataset into 80\% for training and 20\% for validation. The compressor is trained for 10 epochs by Adam optimizer with a batch size of 10 and a learning rate of $1\times 10^{-5}$. After each epoch, we test the model on the validation set and ultimately select the model with the highest validation accuracy. The training process takes approximately 15 hours.

\section{On-Device NPU Deployment Details for LLM}\label{sec:llm_deployment}

For all on-device experiments in this work, we deploy Qwen2.5-3B-Instruct~\citep{yang2024qwen2}, a representative small language model suitable for mobile devices, on a smartphone running Android 15 with 16GB of LPDDR5X memory and 1TB of UFS 4.0 storage, powered by the MT6991 chipset. While we here, as an example, use an up-to-date flagship phone so as to report the state-of-the-art performance values, our proposed scheme is not meant to work on devices of limited models or with a limited range of chipsets only; instead, it is plausible that our proposal naturally extends to a wide range of platforms, whether on cloud or at edge, on high-end smartphones and tablets or on low-cost embedded systems. In the following, we briefly elaborate on the executive details.

\subsection{Model Preparation}

The model is basically an open-source copy downloaded from Hugging Face\footnote{\url{https://huggingface.co/Qwen/Qwen2.5-3B-Instruct}}. Note that the default precision of the model is \texttt{bfloat16}.

\subsection{Conversion and PTQ}

\noindent\textbf{Quantization.} Models to be executed on the mobile NPU must be quantized to fixed-point precisions. Following general practice, we perform PTQ on a CUDA server with int4 weights, int16 activations (4W16A), and int8 KV caches. Rotation matrices~\citep{ashkboos2024quarotoutlierfree4bitinference,liu2024spinquantllmquantizationlearned} are used to improve the accuracy of PTQ.

\noindent\textbf{Shape Fixing.} Contrary to the well-known cloud-based LLM inference that is very flexible in terms of dynamic shapes, batching, etc., the built-in NPU only supports fixed-shape inputs and outputs for inference, so we need to generate separate copies of the model for the encoding and decoding phases, respectively, as they work at different levels of parallelism, essentially using different sizes at certain input dimensions. We mainly consider two dimensions, that is, the sequence length and the context window length. Commonly for mobile deployment, the encoding (a.k.a. prefilling, or prompt) phase employs a sequence length of 128 tokens (which is basically constrained by the maximum level of parallelism in the chip design; sometimes known as the \textit{AR128} mode where AR stands for ``auto-regressive''), gradually filling up the context window with KV caches of multiple input tokens during inference\footnote{Left padding exists if the number of tokens to be encoded in a single inference is not an integer multiple of 128.}. Meanwhile, the sequence length of the decoding phase (a.k.a. generative) model is 1 token (sometimes known as the \textit{AR1} mode), reflecting the nature of auto-regressive generation. For the context length, we generate models with four fixed values, which are 1,024, 2,048, 4,096 and 8,192 tokens, respectively. They would be invoked in each inference according to the total token count, with both input and output tokens included (assumed to be known \textit{a priori}). As a result, we obtain a total of $2\times4=8$ fixed-shape models for both phases and all needed context lengths, whose accuracies we may check offline before proceeding to the compilation stage.

\subsection{Compilation}

\noindent\textbf{Offline Compilation.} Now the models are still not in the NPU's native format. They require compilation into native commands before they can be executed on the NPU. Compilation is typically the process to convert non-native representations of a program, such as high-level coding languages and public model exchange formats, into native commands executable on the bare metal. We execute this step on a PC and reuse the resulting native models in all the runs.

\noindent\textbf{Optimizations.} Note that the only differences between all the 8 fixed-shape models are the input shapes and that all parameters, especially the most space-consuming weights, are shared between them but stored twice for each context length (or totally 8 times in our case), which leads to a substantial waste of the precious on-device storage space. Therefore, it is necessary to extract the shared weights between the two sets of models to reduce the storage footprint. Now, the post-processed files are ready for on-device inference.

\subsection{Inference on Device}

The compiled and optimized models are executed on the smartphone using a llama.cpp~\citep{llama-cpp-2025}-based inference engine modified to support the on-chip NPU. Before we start each experiment, we boost the CPU, memory and NPU, all to their highest possible frequencies, to ensure reproducibility and fair comparisons. Then we simply feed the (decoded) strings before and after they are compressed by the compressor into the on-device models to obtain the corresponding output contents and inference times. Since the input strings are already deemed given at this time, a model of the appropriate context length could easily be selected for each run. We always use the one with the context window just large enough.

\section{On-Device GPU Deployment Details for Compressor}\label{sec:bert_deployment}

To deploy the compressor, we take the same on-device configurations as described in Appendix~\ref{sec:llm_deployment}.  
We first export the model to ONNX\footnote{\url{https://github.com/onnx/onnx}} format (ONNX 1.14 and OPSET 17) and then convert it to a format~\citep{li2024transformer} that can be executed on the smartphone's GPU.
To strike a balance between inference speed and accuracy, we employ mixed precision during inference. Specifically, we use FP32 for the calculation of outlier scores and LayerNorm, while the rest of the model uses FP16 for inference.
Unlike the strict restrictions on fixed input shapes when deploying on NPU, deploying on GPU allows for dynamic input shapes, 
which provides us with great convenience to handle various content lengths from 1,024 to 8,192 tokens.

\begin{figure*}[ph]
  \includegraphics[width=0.95\textwidth]{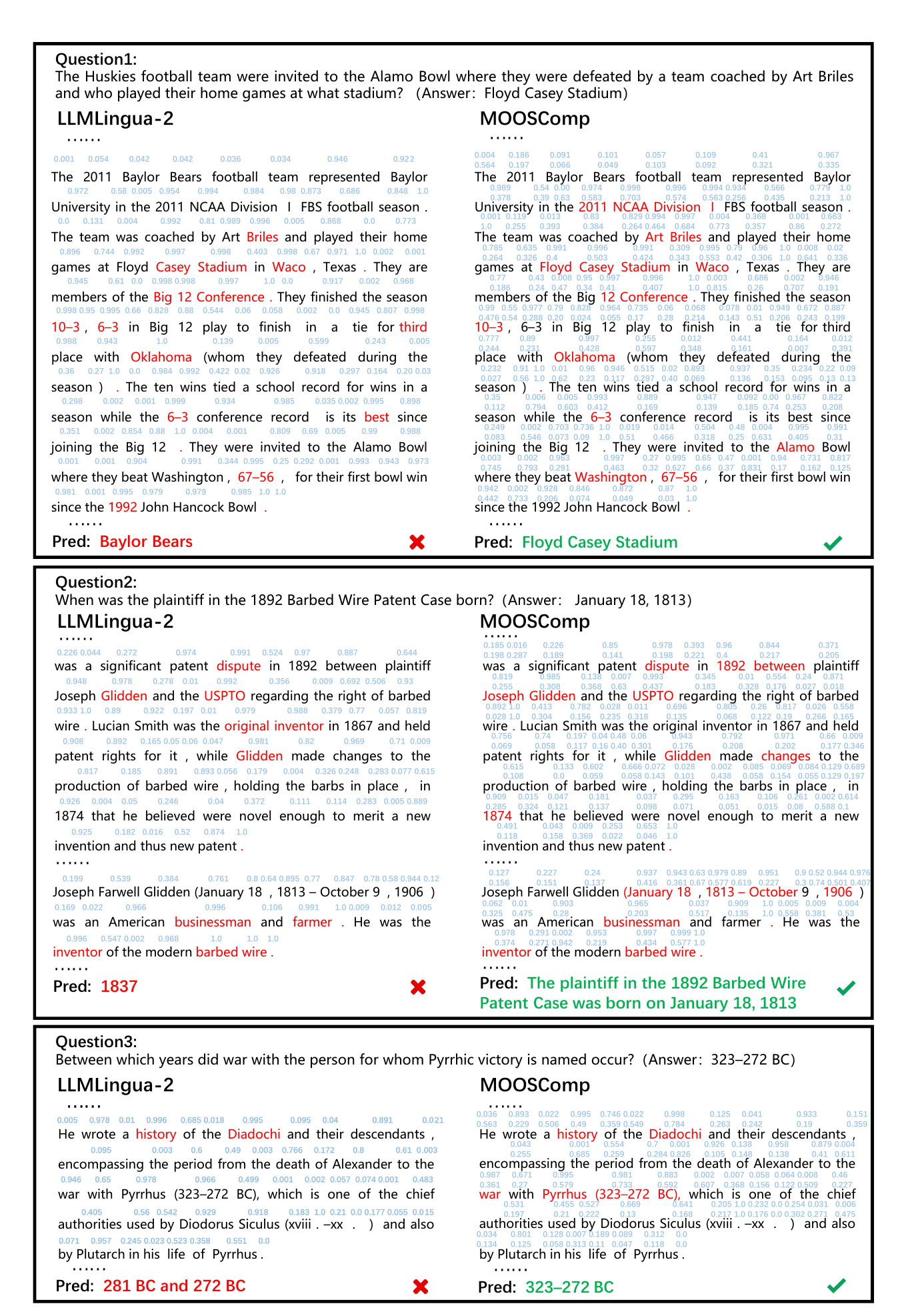}
  \centering
  \caption{Examples of the original and compressed prompts from the LongBench dataset, with the preserved words highlighted in red. \textbf{Left}: Results of LLMLingua-2. The number above each word indicates the probability of that word being preserved. \textbf{Right}: Results of MOOSComp. The two numbers above each word represent the probability of that word being preserved (up) and the outlier score of that word (down), respectively.}
  \label{fig:example1}
\end{figure*}

\section{Examples of Compressed Texts}\label{sec:compressed_texts}

Figure~\ref{fig:example1} illustrates the differences in compression details between MOOSComp and LLMLingua-2. All selected examples are from LongBench.
From Question 1, it is evident that the words ``Alamo Bowl'', ``Art Briles'' and ``Floyd Casey Stadium'' are crucial for answering the question. Although these words have a high probability of being preserved in the results of LLMLingua-2, they are not all retained due to the excessive number of words with similarly high preservation probabilities. This illustrates the issue of over-smoothing, which causes the compressor to lack sufficient distinction between tokens.
On the contrary, our method mitigates this issue through the anti-over-smoothing mechanism, while also allowing for adjustments using the outlier scores.
For answering Question 3, the words "Pyrrhus (323-272 BC)," are important. They are discarded in the results of LLMLingua-2 but retained in the results of MOOSComp. The outlier scores for these words are all quite high. Although the preservation probability of "BC)," is lower than that of "wrote" and "descendants", its outlier score is much higher. The inclusion of the outlier score allows "BC)," to be prioritized for retention over "wrote" and "descendants".

\end{document}